\documentclass[10pt,twocolumn,letterpaper]{article}

\pdfoutput = 1

\usepackage{iccp}
\usepackage{times}
\usepackage{epsfig}
\usepackage{graphicx}
\usepackage{amsmath}
\usepackage{amssymb}
\usepackage{url}
\usepackage{enumitem}
\usepackage{multirow}
\usepackage{cite}

\iccpfinalcopy 

\ificcpfinal\fi

\newcommand{\bfI}{{\bf I}}
\newcommand{\bfl}{{\bf l}}
\newcommand{\bfv}{{\bf v}}
\newcommand{\bfp}{{\bf p}}
\newcommand{\bfn}{{\bf n}}
\newcommand{\bfs}{{\bf s}}
\newcommand{\bfc}{{\bf c}}

\newcommand{\reals}{{\mathbb R}}

\begin{document}

\title{A Dictionary-based Approach for Estimating Shape and \\ Spatially-Varying Reflectance}
\author{Zhuo Hui and Aswin C.~Sankaranarayanan\thanks{The authors were supported in part by the NSF grant CCF-1117939. Email:  {\tt\small \{zhui, saswin\}@andrew.cmu.edu}}\\
ECE Department, Carnegie Mellon University, Pittsburgh, PA}

\maketitle

\thispagestyle{empty}

\begin{abstract}
We present a technique for estimating the shape and reflectance of an object in terms of its surface normals and spatially-varying BRDF.
We assume that multiple images of the object are obtained under fixed view-point and varying illumination, i.e, the setting of photometric stereo.
%
Assuming that the BRDF at each pixel lies in the non-negative span of a known BRDF dictionary, 
we derive a per-pixel surface normal and BRDF estimation framework that requires neither iterative optimization techniques nor careful initialization, both of which are endemic to most state-of-the-art techniques.
We showcase the performance of our technique on a wide range of simulated and real scenes where we outperform competing methods.
	
\end{abstract}


\section{Introduction} \label{sec:intro}
Photometric stereo \cite{woodham1980photometric} seeks to estimate the shape of an object from images which are obtained from a static camera and under varying lighting.
While there has been remarkable progress in photometric stereo, the vast majority of techniques are devoted to scenes that exhibit simple reflectance properties.
%
%
Yet, this creates a significant disconnect between theory and practice since the vast majority of real-life scenes involve materials with complex reflectance properties that interact with light in myriad number of ways.
\begin{figure}
\center
	\includegraphics[width=.475\textwidth]{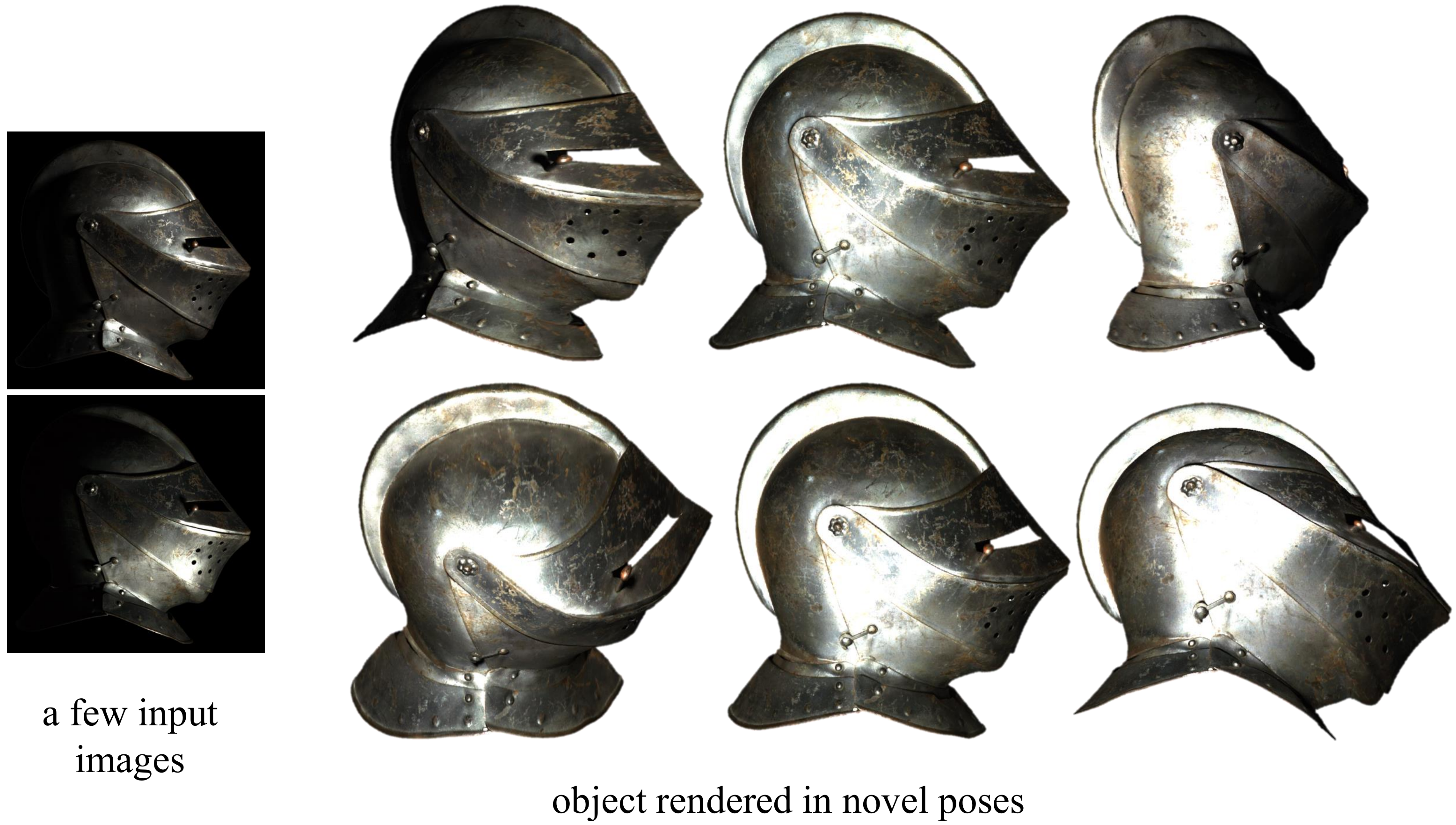}
	\caption{{\bf Recovery of surface normals and spatially-varying BRDF.} We propose a framework for per-pixel estimation of surface normal and BRDF in the setting of photometric stereo. Shown above are the estimated shape and rendered images of a visually-complex object. The results were obtained from 250 images.}
	\label{fig: teaser}
\end{figure}

In this paper, we present a method for recovering the surface normals and the reflectance of opaque objects with complex spatially-varying reflectance.
The key challenge here is that the reflectance, characterized in terms of its spatially-varying bidirectional reflectance distribution function (SV-BRDF), and the shape, characterized in terms of surface normals, are coupled and need to be jointly estimated.
Further, the SV-BRDF is a very high-dimensional signal and, in the absence of additional assumptions, requires a large number of input images for robust estimation.

A common assumption for computational tractability is that the BRDF at each pixel is a weighted combination of a \textit{few} reference BRDFs \cite{lawrence2006inverse}.
We now need to estimate only the reference BRDFs and their abundances at each pixel which is a significant reduction in the dimensionality of the unknowns.
In Goldman et al.\ \cite{goldman2010shape}, the isotropic Ward model, a parametric model is used to characterize the reference BRDFs.
Alldrin et al.\ \cite{alldrin2008photometric} assume that the reference BRDFs are approximated by the so-called bivariate model, a non-parametric model that approximates the 4D BRDF as a 2D signal.
In both cases, the problem of shape and SV-BRDF estimation reduces to alternating minimization over the surface normals, the reference BRDFs, and abundances of the reference BRDFs at each pixel.
The drawback of these approaches is that the optimization is not just computationally expensive but also has a critical dependence on the ability to find a good initial solution since the underlying problem is non-convex and riddled with local minima.

An alternate approach called example-based photometric stereo \cite{hertzmann2005example,ren2011pocket} introduces reference objects --- objects with known shape --- in the scene.
These techniques rely on the concept of \textit{orientation consistency} \cite{hertzmann2005example} which suggests that two surface elements with identical normal and BRDF will take the same appearance when placed in the same illumination.
If the reference object had the same BRDF as the target object, we can obtain the shape of the target simply by comparing the intensity profile observed at a pixel on the target to those observed on the reference object. 
When the target's BRDF is spatially-varying, it can be shown that two reference objects --- one diffuse and the one specular --- are sufficient to recover the surface normals of the target by approximating the unknown BRDF at each pixel as a non-negative linear combination of the reference BRDFs \cite{hertzmann2005example}.
While introducing reference objects is not always desirable, example-based photometric stereo produces precise shape estimates without requiring the knowledge of lighting. 

\begin{figure}[!ttt]
\center
\includegraphics[width=0.475\textwidth]{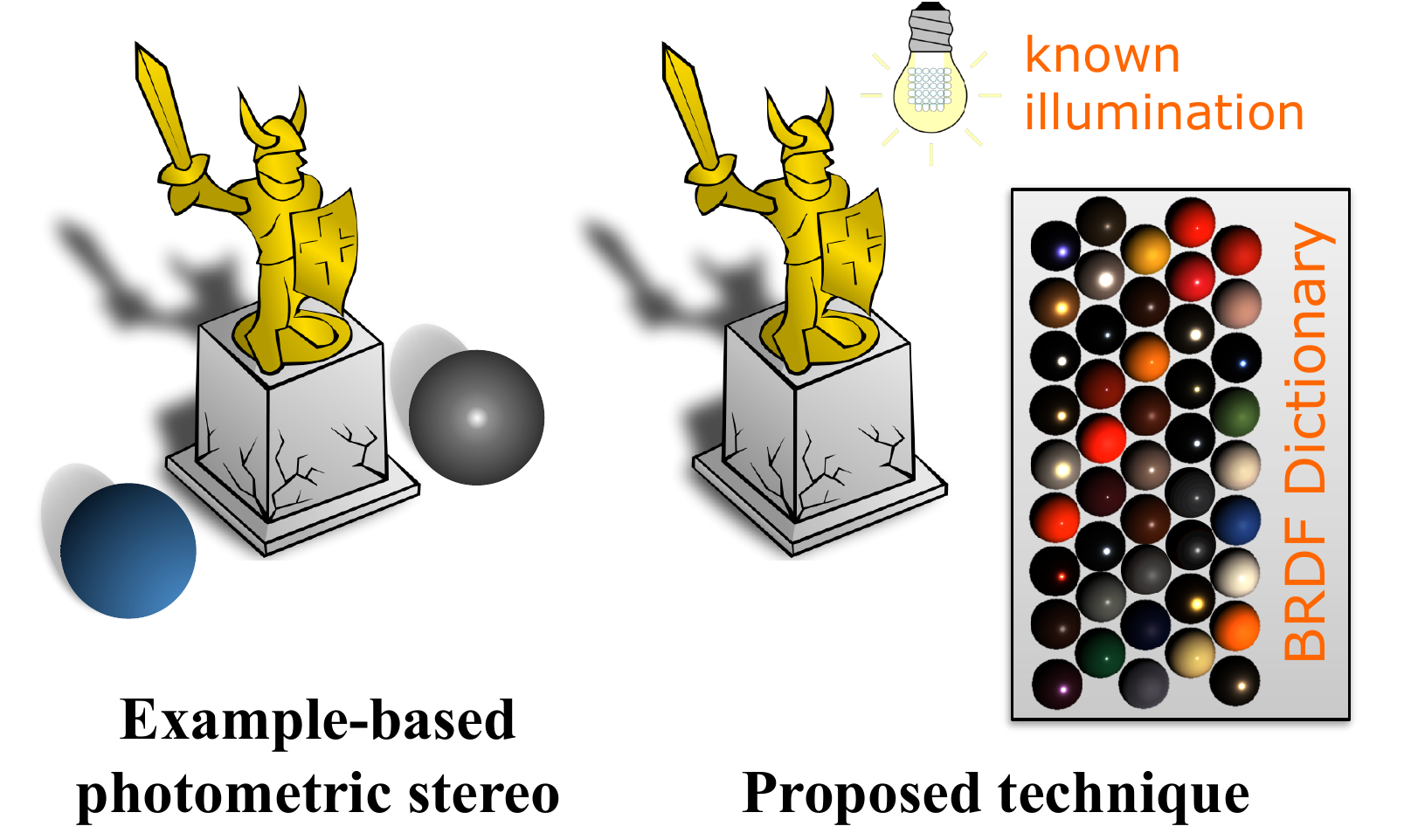}
\vspace{-7mm}
\caption{{\bf Virtual examples.} In example-based photometric stereo \cite{hertzmann2005example}, objects with known shape and reflectance are introduced into a scene. In contrast, we use a dictionary of BRDFs to render ``virtual examples'' that guide the normal estimation problem. This enables us to handle scenes with very complex reflectance, since we can use a larger collection of virtual examples.}

\label{fig:example}
\end{figure}

The technique proposed in this paper relies on the core principle of example-based photometric stereo {\em without actually introducing reference objects into the scene}. 
Instead, given a dictionary whose atoms are BRDFs associated with a wide range of materials, 
we can render virtual spheres, one for each atom in the dictionary, under the knowledge of the scene illumination (typically a distant point light source).
This provides a set of ``virtual examples'' that can be used to obtain a per-pixel estimate of the shape and reflectance of the scene with arbitrary spatially-varying BRDF  (see Figure \ref{fig:example}).
The assumption that we make is that the unknown BRDF at each pixel lies in the non-negative span of the dictionary atoms.
%
%
We show that the surface normals and the BRDFs can be estimated via a sequence of tractable linear inverse problems.
This obviates the need for complex iterative optimization techniques as well as careful initialization required to avoid convergence to local minima.
The interplay of these ideas leads to a robust surface normal and SV-BRDF estimation technique that provides state-of-the-art results on challenging real-life scenes (see Figure \ref{fig: teaser}).

\vspace{-1mm}
\paragraph{Contributions.}
\vspace{-2mm}
We make the following contributions.
\begin{description}[leftmargin=*]
  \setlength\itemsep{0em}
\item [][\textbf{Model}] We propose the use of a dictionary of BRDFs to regularize the surface normal and SV-BRDF estimation. The BRDF at each pixel of an object is assumed to lie in the non-negative span of the dictionary atoms. 

\item [][\textbf{Normal estimation}] We show that the surface normal at each pixel can be efficiently estimated using a coarse-to-fine search.

\item [][\textbf{SV-BRDF estimation}] Given the surface normals, we recover the BRDF at each pixel independently by solving a linear inverse problem that enforces sparsity in the occurrence of the reference BRDFs at the pixel.

\item [][\textbf{Validation}] We showcase the accuracy of the shape and SV-BRDF estimation technique on a wide range of simulated and real scenes and demonstrate that the proposed technique outperforms state-of-the-art.
\end{description}

\section{Prior work} \label{sec:prior}
%
%
%
In this section, we review some of the key techniques for non-Lambertian shape estimation.

\vspace{-3mm}
\paragraph{The diffuse + specular model.} 
It is well known that the collections of images of a convex Lambertian object typically lies close to a low-dimensional subspace \cite{basri2003lambertian,ramamoorthi2002analytic}.
This naturally leads to techniques \cite{wu2011robust,ikehata2012robust,yu2010photometric,yu2013outdoor} that robustly fit a low-dimensional subspace, capturing the Lambertian component while isolating non-Lambertian components, such as specularities, as sparse outliers.
However, these techniques have restrictive assumptions on the range of BRDFs to which they are applicable, and more importantly, miss out on powerful cues to the shape of the object that is often present in specular highlights.

\vspace{-4mm}
\paragraph{Parametric BRDFs.}
%
Parametric models such as the Blinn-Phong \cite{blinn1976texture}, Ward \cite{ward1992measuring}, Oren-Nayar \cite{oren1995generalization}, and Cook-Torrance model \cite{cook1982reflectance} are based on macro-behavior established using specific micro-facet models on the materials, and have been widely used in computer graphics.
In the context of shape and SV-BRDF estimation, Goldman et al.\ \cite{goldman2010shape} utilize the isotropic Ward model \cite{ward1992measuring} to reduce the dimensionality of the inverse problem.
Oxholm and Nishino \cite{oxholm2012shape,oxholmmultiview} further extend this idea by introducing a probabilistic formulation to estimate the BRDFs from a single image under natural lighting conditions.  
However, parametric models are inherently limited in their ability to provide precise approximations to the true BRDFs and further, lead to challenging and ill-conditioned optimization problems.
   
\vspace{-4mm}
\paragraph{Isotropic BRDFs.}
Isotropic materials exhibit a form of symmetry, wherein the reflectance of the material is unchanged when the incident and outgoing directions are jointly rotated about the surface normal.
This enables the representation of isotropic BRDFs as the function over three as opposed to four angles.
In the context of photometric stereo, Alldrin and Kriegman \cite{alldrin2007toward} observe that, for isotropic materials, the surface normal at each point can be restricted to lie on a plane.
When the isotropic BRDFs has a single dominant lobe, Shi et al.\ \cite{shi2012elevation} resolve the planar ambiguity and show that the surface normals can be uniquely determined.
Higo et al.\ \cite{higo2010consensus} utilize properties of isotropy, visibility and monotonicity to restrict the solution space of the surface normal at each pixel. This enables a framework for shape estimation without the need for radiometric calibration.
Finally, a bivariate approximation for isotropic materials is used in Romeiro et al.\ \cite{romeiro2008passive,romeiro2010blind} to estimate the BRDF of a known shape from a single image and without knowledge of the scene illumination.

\vspace{-4mm}
\paragraph{Reference basis model.}
As mentioned in the introduction, a common assumption for scenes with SV-BRDF is that the per-pixel BRDF is generated from a few unknown reference BRDFs \cite{lawrence2006inverse,goldman2010shape,alldrin2008photometric,chandraker2011image}.
Invariably, this leads to a multi-linear optimization in high-dimensional variables (the reference BRDFs) that is highly dependent on initial conditions.
In contrast, our proposed technique avoids the need to estimate high-dimensional optimization by evoking knowledge of a dictionary of BRDFs.

\section{Problem setup} \label{sec:overview}

\paragraph{Setup.}
We make the following assumptions.
First, the camera is orthographic and hence, the viewing direction $\bfv \in {\mathbb R}^3$ is constant across all scene points.
Second, the scene illumination is assumed to be from a distant point light source.
The light sources are assumed to be of constant brightness (equivalently, that calibration is known) and their direction is known. We denote $\bfl_k \in {\mathbb R}^3$ to refer to the lighting direction in the $k$-th image $I^k$.
For a light-stage, this information is typically obtained by a one-off calibration.
Third, the effects of long-range illumination such as cast shadows and inter-reflections are assumed to be negligible; this is satisfied for objects with a convex shape.
Finally, the radiometric response of the camera is assumed to be known.


\vspace{-4mm}
\paragraph{BRDF representation.} 
We follow the isotropic BRDF representation used in \cite{rusinkiewicz1998new} in which a three-angle coordinate system based on half angles is used.
Specifically, the BRDF is expressed as a function $\rho(\theta_h, \theta_d, \phi_d)$ 
with $\theta_h, \theta_d \in [0, \pi/2)$ and $\phi_d \in [0, 2\pi)$.
However, by Helmholtz's reciprocity, the BRDF exhibits the following symmetry: $\rho(\theta_h, \theta_d, \phi_d) = \rho(\theta_h, \theta_d, \phi_d + \pi),$ and hence it is sufficient to express $\phi_d \in [0, \pi)$.
Following \cite{matusik2003data}, we use a $1^\circ$ sampling of each angle.
As a consequence, a BRDF is represented as a point in a $T = 90 \times 90 \times 180  =  1,458,000$-dimensional space.
When we deal with color images, we have a BRDF for each color channel and hence, the dimensionality of the BRDF goes up proportionally.

Consider a scene element with BRDF $\rho \in \reals^T$, surface normal $\bfn$, illuminated from a point light source from a direction $\bfl$ and viewed from a direction $\bfv$.
For this configuration of normal, incident light and viewing direction, the BRDF value is simply a linear functional of the vector $\rho$:
\[ \bfs_{\{\bfl, \bfv; \bfn\}}^\top \rho, \]
where $\bfs_{\{\bfl, \bfv; \bfn\}}$ is a vector that encodes the geometry of the configuration.
In essence, the vector samples the appropriate entry from $\rho$, allowing for the appropriate interpolation if the required value is off the sampling-grid.

\begin{figure*}[!ttt]
	\center
	\includegraphics[width=\textwidth]{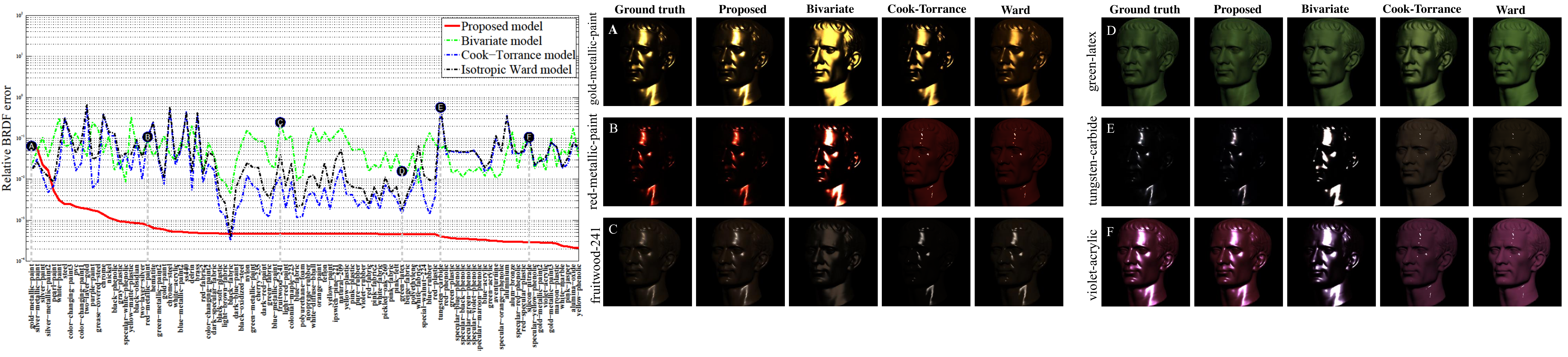}
	\caption{\textbf{Accuracy of BRDF models on the MERL database \cite{matusik2003data}.} For the $100$ materials in the database, we plot the approximation accuracy in relative RMS error \cite{ngan2005experimental} (also see (\ref{eqn:relBRDF})) for the proposed,  bivariate \cite{romeiro2008passive},  Cook-Torrance \cite{cook1982reflectance},  and the Ward  \cite{ward1992measuring} models. For the proposed model, we use a leave-one-out scheme, wherein for each BRDF the remaining $99$ BRDFs in the database are used to form the dictionary. The proposed model outperforms competing models both quantitatively (left) as well as in visual perception (right).}
	\label{fig:brdf_model}
\end{figure*}

\vspace{-4mm}
\paragraph{Problem formulation.}
Our goal is to recover the surface normals and the SV-BRDF in the context of photometric stereo; i.e,  multiple images of an object $\{I^1, \ldots, I^Q\}$ obtained from a static camera under varying lighting.
The intensity value $I^i_{\bfp}$ observed at pixel $\bfp = (x,y)$ with lighting $\bfl_i$ can be written as
\begin{equation}
I^i_{\bfp} = (\bfs_{\{\bfl_i, \bfv; \bfn_\bfp\}}^\top  \rho_\bfp ) \cdot \max \{ 0, \bfn_\bfp^\top \bfl_i \},
\label{eqn:problem}
\end{equation}
where $\rho_\bfp$ is the BRDF and $\bfn_\bfp$ is the surface normal at pixel $\bfp$, respectively, and $\max \{ 0, \bfn_\bfp^\top \bfl_i \}$ accounts for shading. 
%
%

Given multiple intensity values at pixel $\bfp$, one for each lighting direction $\{\bfl_1, 
\ldots, \bfl_Q\}$, we can write 
\begin{eqnarray}
\bfI_\bfp &=& \left( \begin{array}{c} I^1_\bfp \\ \vdots \\ I^Q_\bfp \end{array} \right)  = \left[     \begin{array}{c} 
 \max \{ 0, \bfn_\bfp^\top \bfl_1 \} \cdot \bfs_{\{\bfl_1, \bfv; \bfn_\bfp\}}^\top \\
 \vdots \\
 \max \{ 0, \bfn_\bfp^\top \bfl_Q \} \cdot \bfs_{\{\bfl_Q, \bfv; \bfn_\bfp\}}^\top \\
 \end{array} \right] \rho_\bfp \nonumber\\
 &=& A(\bfn_\bfp) \rho_\bfp
\label{eq:defeq}
\end{eqnarray}
Given the intensities, $\bfI_\bfp$, observed at a pixel $\bfp$ and knowledge of lighting directions $\{\bfl_1, \ldots, \bfl_Q\}$, we seek to estimate the surface normal $\bfn_\bfp$ and the BRDF $\rho_\bfp$ at the pixel.
This problem is intractable without additional assumptions that constrain the BRDF to a lower-dimensional space.

\vspace{-4mm}
\paragraph{Model for BRDF.} The key assumption that we make is that the BRDF at a pixel $\bfp$ lies on the non-negative span of the atoms of a BRDF dictionary. 
Specifically, given dictionary $D = [ \rho^1, \rho^2, \cdots, \rho^M ]$, we assume that the BRDF at pixel $\bfp$ can be written as
\[ \rho_\bfp = D \bfc_\bfp, \quad \bfc_\bfp \ge 0, \]
where $\bfc_\bfp \in \reals^M$ are the abundances of the dictionary atoms.

In essence, we have now constrained the BRDF to lie in an $M$-dimensional cone.\footnote{A more appropriate model for the BRDF is that $(D \bfc) \ge 0.$ However, this leads to significantly higher-dimensional constraints. We instead use a sufficient condition to achieve this, $\bfc \ge 0$.}
This provides immense reduction in the dimensionality of the unknowns at the expense of introducing a model misfit error.
Indeed the success of this model relies on having a dictionary that is sufficiently rich to cover a wide range of interesting materials.
Figure \ref{fig:brdf_model} shows the accuracy of various BRDF models on the MERL BRDF database \cite{matusik2003data}.

We also assume that $\bfc_\bfp$ is sparse, suggesting that BRDF at each pixel is the linear combination of a \textit{few} dictionary atoms.
The sparsity constraint avoids over-fitting to the intensity measurements $\bfI_\bfp$ as well as provides a regularization for under-determined problems.

\vspace{-4mm}
\paragraph{Solution outline.} An estimate of the surface normal and BRDF at a pixel $\bfp$ can be obtained by solving
\begin{equation}
\begin{array}{r}
\{ \widehat{\bfn}_\bfp, \widehat{\bfc}_\bfp \}  = \underset{\bfn, \bfc}{\arg\min}  \, \|  \bfI_\bfp - A(\bfn) D \bfc \|_2^2 + \lambda \| \bfc \|_1 \\
\textrm{s.t}  \quad \bfc \ge 0, \| \bfn\|_2 = 1
\end{array}
\label{eq:daeq}
\end{equation}
The $\ell_1$-penalty serves to enforce sparse solutions, with $\lambda \ge 0$ determining the level of sparsity in the solution.
The optimization problem in (\ref{eq:daeq}) is non-convex due to unit-norm constraint on the surface normal $\bfn$ as well as the term $A(\bfn) D \bfc$.
Our solution methodology consists of two steps:
(i) \textit{Surface normal estimation.} We perform an efficient multi-scale search that provides us with a precise estimate of the surface normal at pixel $\bfp$ (see Section \ref{sec:Normal}); and, 
(ii) \textit{BRDF estimation.} We solve (\ref{eq:daeq}) only over $\bfc$ with the normal fixed to obtain the BRDF at $\bfp$ (see Section \ref{sec:BRDF}).

\section{Surface normal estimation} \label{sec:Normal}
In this section, we describe an efficient per-pixel surface normal estimation algorithm.

\subsection{Virtual example-based normal estimation}
Our surface normal estimation is an extension of the method proposed in \cite{hertzmann2005example}, where two spheres --- one diffuse and one specular --- are introduced in a scene along with the target object.
To obtain the surface normals at an pixel $\bfp$ on the target, the intensity observed at pixel $\bfp$, $\bfI_\bfp$, is matched to those on the reference spheres.
The reference spheres provide a sampling of the space of the normals and hence, we can simply treat them as a collection of candidate normals $\mathcal{N}$.
By orientation consistency, the surface normal estimation now reduces to finding the candidate normal that can best explain the intensity profile $\bfI_\bfp$.
Given a candidate normal $\widetilde{\bfn}$, we have two intensity profiles, $\bfI_D(\widetilde{\bfn})$ and $\bfI_S(\widetilde{\bfn})$, one each for the diffuse and specular sphere, respectively.
The estimate of the surface normal at pixel $\bfp$ is given as
\[ \widehat{\bfn}_\bfp = \underset{\widetilde{\bfn} \in \mathcal{N}}{\arg \min} \underset{ a_1, a_2 \ge 0}{\min} \| \bfI_\bfp - a_1 \bfI_D(\widetilde{\bfn}) -a_2 \bfI_S(\widetilde{\bfn}) \|. \]
In \cite{hertzmann2005example}, this is solved by scanning over all the pixels/candidate normals on the reference spheres.

\vspace{-4mm}
\paragraph{Rendering virtual spheres.}
We rely on the same approach as \cite{hertzmann2005example} with the key difference that we virtually render the reference spheres.
The virtual spheres are rendered as follows.
Given the lighting directions $\{ \bfl_1, \ldots, \bfl_Q \}$ and the BRDF dictionary $D = [ \rho^1, \ldots, \rho^M]$, for each candidate normal $\widetilde{\bfn} \in \mathcal{N}$, we render a matrix $B(\widetilde{\bfn}) = [ b_{ij}(\widetilde{\bfn}) ] \in \reals^{Q \times M}$ such that $b_{ij}(\widetilde{\bfn})$ is the intensity observed at a surface with normal $\widetilde{\bfn}$ and BRDF $\rho^j$, under lighting $\bfl_i$. 
\[ b_{ij}(\widetilde{\bfn}) = \max \{ 0, \widetilde{\bfn}^\top \bfl_i \} \cdot \bfs_{\{\bfl_i, \bfv; \widetilde{\bfn}\}}^\top  \rho^j, \]
%
%
We render one such matrix $B(\cdot)$ for each candidate normal in $\mathcal{N}$.
Given these virtually rendered spheres, we can solve (\ref{eq:daeq}) by searching over all candidate normals.
%
%

\vspace{-4mm}
\paragraph{Brute-force search.}
For computationally efficiency, we drop the sparsity constraint in (\ref{eq:daeq}). 
We empirically observed that dropping the sparsity constraint made little difference in the estimated surface normals.
Now, given the intensity profile $\bfI_\bfp$ at pixel $\bfp$ and noting that $B(\widetilde{\bfn}) = A(\widetilde{\bfn}) D$, solving (\ref{eq:daeq}) reduces to:
\begin{equation}
 \widehat{\bfn}_\bfp = \underset{\widetilde{\bfn} \in \mathcal{N}}{\arg\min} \quad \underset{\bfc \ge 0}{\min}\ \ \| {\bf I}_\bfp - B(\widetilde{\bfn}) \bfc \|.
\label{eq:normalest}
\end{equation}     
The unit-norm constraint on the surface normals is absorbed into the candidate normals being unit-norm.
The optimization problem in (\ref{eq:normalest}) requires solving a set of a non-negative least squares (NNLS) sub-problems, one for each element of $\mathcal{N}$.
For the results in the paper, we used the {\tt lsnonneg} function in MATLAB to solve the NNLS sub-problems.

The accuracy and the computational cost in solving (\ref{eq:normalest}) depends solely on the cardinality of the candidate set $\mathcal{N}$, $|\mathcal{N}|$.
We obtain $\mathcal{N}$ by uniform or equi-angular sampling on the sphere \cite{harman2010decompositional}.
%
As a consequence, the accuracy of the normal estimates, on an average, cannot be better than the half the angular spacing of the candidate set.
Yet, the smaller the angular spacing of $\mathcal{N}$, the larger is its cardinality.
For example, a $5^\circ$ equi-angular sampling over the hemisphere requires approximately 250 candidates while a $0.5^\circ$ requires 20000 candidates.
Given that the time-complexity of the brute-force search is linear in $|\mathcal{N}|$, the computational costs for obtaining very precise normal estimates can be over-whelming.
To alleviate this, we outline a coarse-to-fine search strategy that is remarkably faster than the brute-force approach with little loss in accuracy.

\subsection{Coarse-to-fine search}

\begin{figure}[!ttt]
	\center
	\includegraphics[width=0.475\textwidth]{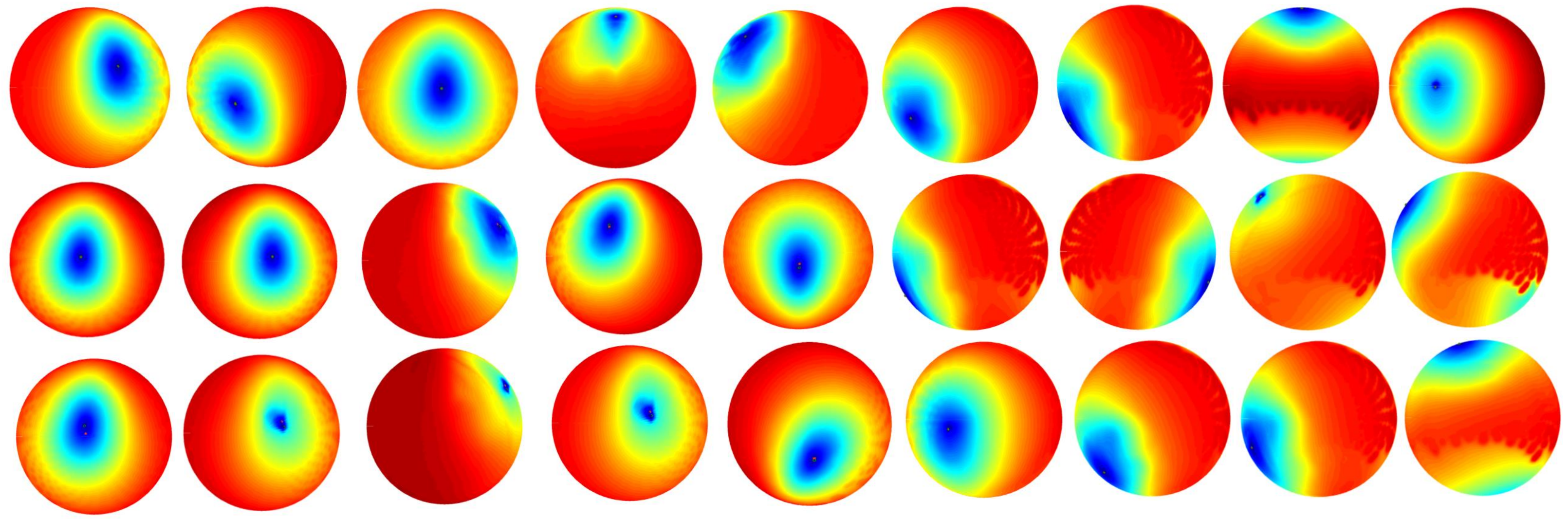}
	\caption{\textbf{The error as a function of candidate normals for a few test examples}. We can observe that the global minima is compact and the error increases largely monotonically in its vicinity. This motivates our coarse-to-fine search strategy.}
	\label{fig:normal_error}
\end{figure}

Figure \ref{fig:normal_error} shows the value of $$E(\widetilde{\bfn}) = \min_{\bfc \ge 0} \| \bfI_\bfp - B(\widetilde{\bfn}) \bfc \|$$ as a function of the candidate normal $\widetilde{\bfn}$ for a few examples.
In our simulations, we observed that there is a gradual increase in error value as we moved away from the global minima of $E(\cdot)$.
We exploit this to design a coarse-to-fine search strategy where we first evaluate the candidate normals at a coarse sampling and subsequently search in the vicinity of this solution but at a finer sampling.

Specifically, let $\mathcal{N}_\theta$ be the set of equi-angular sampling on the unit-sphere where the angular spacing is $\theta$ degrees.
Given a candidate normal $\widetilde{\bfn}$, we define 
\[ C_\theta(\widetilde{\bfn}) = \{ \bfn\ |\ \langle \bfn, \widetilde{\bfn} \rangle \ge \cos \theta,\ \| \bfn \|_2 = 1 \} \]
as the set of unit-norm vectors within $\theta$-degrees from $\widetilde{\bfn}$,

In the first iteration, we initialize the candidate normal set $\mathcal{N}^{(1)} = \mathcal{N}_{\theta_1}$.
Now, at the $j$-th iteration, we solve (\ref{eq:normalest}) over a candidate set $\mathcal{N}^{(j)}$.
Suppose that $\widehat{\bfn}^{(j)}$ is the candidate normal where the minimum occurs at the $j$-th iteration.
The candidate set for the $(j+1)$-th iteration is constructed as 
\[ j \ge 1, \quad \mathcal{N}^{(j+1)} = C_{\theta_j}(\widehat{\bfn}^{(j)}) \cap \mathcal{N}_{\theta_{j+1}}, \]
with $\theta_{j+1} < \theta_j$.
That is, the candidate set is simply the set of all candidates at a finer angular sampling that are no greater than the current angular sampling from the current estimate.
This is repeated till we reach the finest resolution at which we have candidate normals.
For the results in this paper, we use the following values: $\theta_1 = 10^\circ, \theta_2 = 5^\circ, \theta_3 = 3^\circ, \theta_4 = 1^\circ,$ and $\theta_5 = 0.5^\circ$. 
For efficient implementation, we pre-render $B(\widetilde{\bfn})$ for $\widetilde{\bfn} \in \mathcal{N}_{\theta_1} \cup  \cdots \cup \mathcal{N}_{\theta_5}$.  

The computational gains obtained via this coarse-to-fine sampling strategy are immense. 
Table \ref{tab:cmp} shows the run-time and precision of both brute force and coarse-to-fine normal estimation strategy for different levels of angular sampling in the generation of the candidate normal set. 
As expected the run-time of the brute force algorithm is linear in the number of candidates.
In contrast, the coarse-to-fine strategy requires a tiny fraction of this time while nearly achieving the same precision as the brute force strategy.

\begin{table}[!ttt]
	\center
	\includegraphics[width=0.475\textwidth]{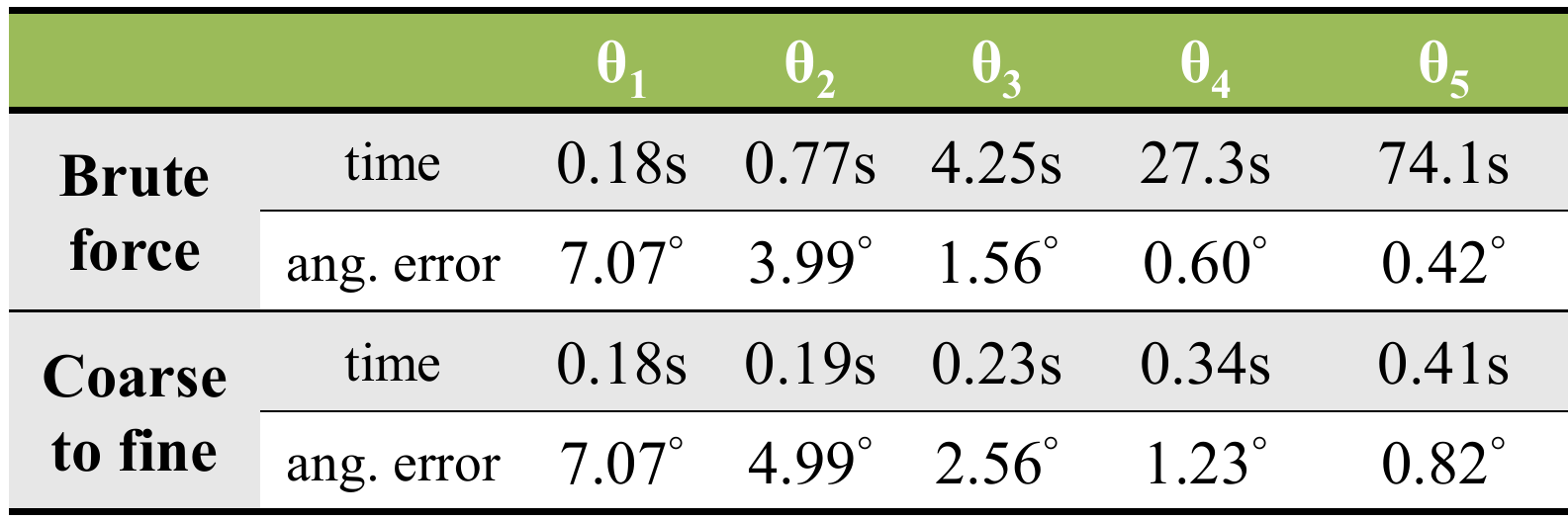}
	\caption{Comparison of brute-force and coarse-to-fine normal estimation for different angular samplings in the candidate normals. Shown are aggregate statistics over 100 randomly generated trials.}
	\label{tab:cmp}
\end{table}

%


While the solution to (\ref{eq:normalest}) also produces an estimate of the BRDF at the pixel, this estimate is often poor due to lack of the sparse-regularizer that serves to avoid over-fitting to the observed intensities.
In the next section, we use this normal estimate to obtain a per-pixel BRDF estimate.

\section{Reflectance estimation} \label{sec:BRDF}
Given the surface normal estimate $\widehat{\bfn}_\bfp$, we obtain an estimate of the BRDF at each pixel, individually, by solving 
\begin{align}
	\widehat{\bfc}_\bfp\ &= \underset{{\bfc \ge 0} }{\arg\min}\  \| {\bf I}_\bfp - B(\widehat{\bfn}_\bfp) \bfc \|^2_2 + \lambda \|\bfc\|_1.
	\label{eqn: brdfest}
\end{align} 
%
%
%
The use of the $\ell_1$-regularizer promotes sparse solutions and primarily helps in avoiding over-fitting to the observed intensities.
The optimization problem in (\ref{eqn: brdfest}) is convex and we used CVX \cite{grant2008cvx}, a general purpose convex solver, to obtain solutions.
The estimate of the BRDF at pixel $\bfp$ is given as $\widehat{\rho}_\bfp = D \widehat{\bfc}_\bfp$.
The value of $\lambda$ was manually tuned for best performance. 
For color-imagery, we solve for the coefficients associated with each color channel separately.

When we know a priori that multiple pixels share the same BRDF, then we can solve (\ref{eqn: brdfest}) simply by concatenating their corresponding intensity profiles and their respective $B(\cdot)$ matrices.
As is to be expected, pooling intensities observed at multiple pixels significantly improves the quality of the estimates.
Yet, while spatial averaging or spatial priors improve the quality of the estimate, inherently they require the object to exhibit smooth spatial-variations in its BRDF.
The advantage of our per-pixel BRDF estimation framework is the ability to handle arbitrarily complex spatial variations in the BRDF at cost of noisier estimates.
In the next section, we carefully characterize the performance of our proposition using synthetic and real examples.

\section{Results} \label{sec:results}
We characterize the performance of our technique using both synthetic and real datasets.

\subsection{Synthetic experiments}
We use the BRDFs in the MERL database \cite{matusik2003data} in a leave-one-out scheme for testing the accuracy of our proposed algorithms for surface normal and BRDF estimation.
Specifically, when we simulate a test object using a particular material, the dictionary is comprised of BRDFs of the remaining $M=99$ materials from the database.
We used the configuration in the light-stage described in \cite{einarsson2006relighting} for our collection of lighting directions.

\begin{figure}[!ttt]
	\centering
	\includegraphics[width=.475\textwidth]{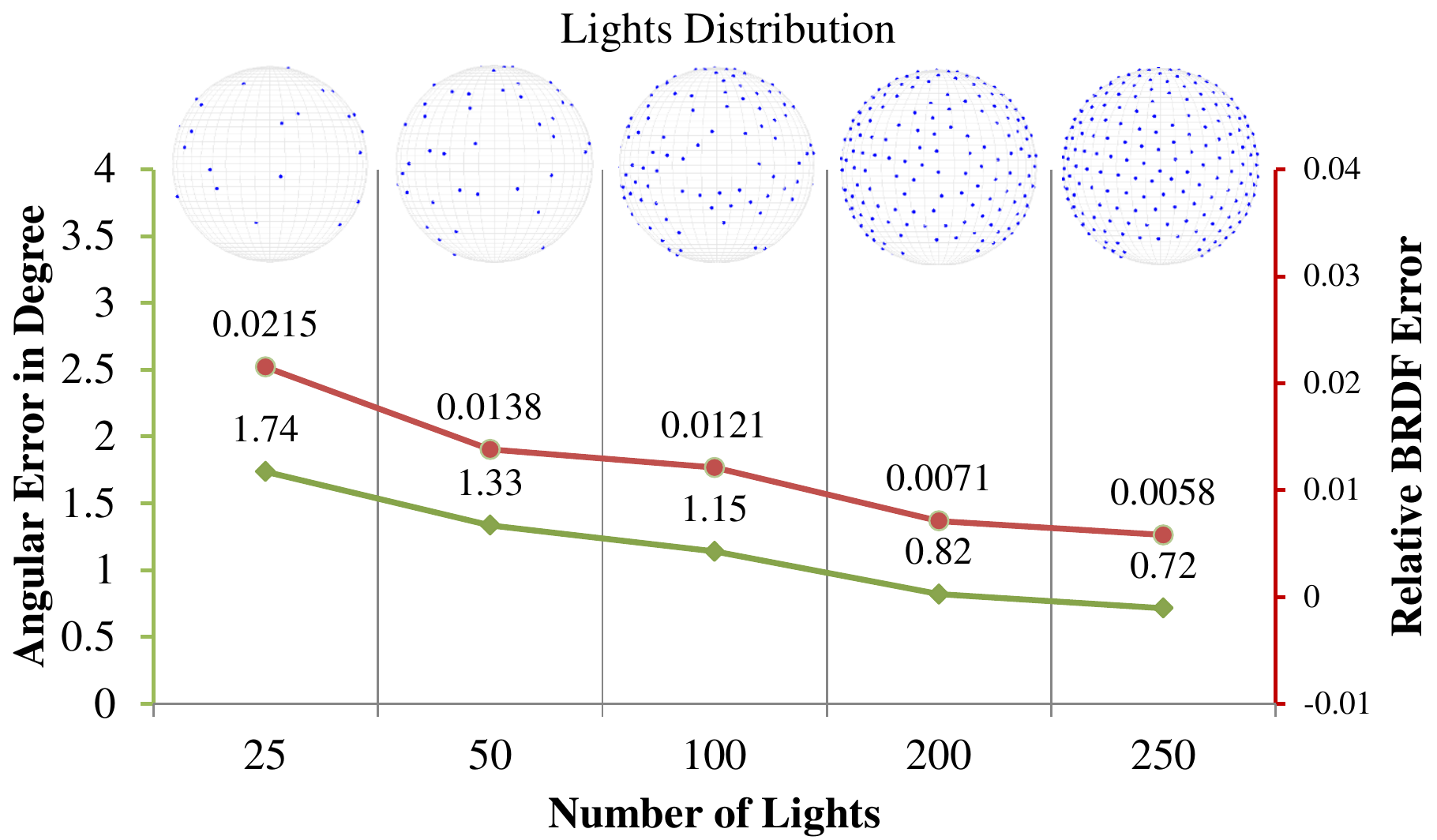}
	\caption{ \textbf{Normal and BRDF estimation with varying number of images.} Given an input number of images, the angular errors (in green) and relative BRDF errors (in red) were obtained by averaging across all 100 BRDFs and across 20,000 randomly-generated normals per material.}
	\label{fig:lightnumber_err}
\end{figure}

\vspace{-3mm}
\paragraph{Varying number of images.} 
Figure \ref{fig:lightnumber_err} characterizes the errors in surface normal and BRDF estimation for varying number of input images or equivalently, lighting directions.
We report the average error computed by randomly generating 20,000 normals per material and varying across all $100$ material BRDFs in the database.
This experiment is similar in setup to the one reported in \cite{shi2012elevation} which, to our knowledge, is one of the most accurate techniques for photometric stereo on isotropic BRDFs.
In \cite{shi2012elevation}, for 200 images, the error in estimating the elevation angle \textit{when the azimuth is known} is reported as $0.88^\circ$; in contrast, the proposed technique has an error of $0.82^\circ$ in estimating the surface normal without any prior knowledge of the azimuth.

\begin{figure}[!ttt]
\center
	\includegraphics[width=0.475\textwidth]{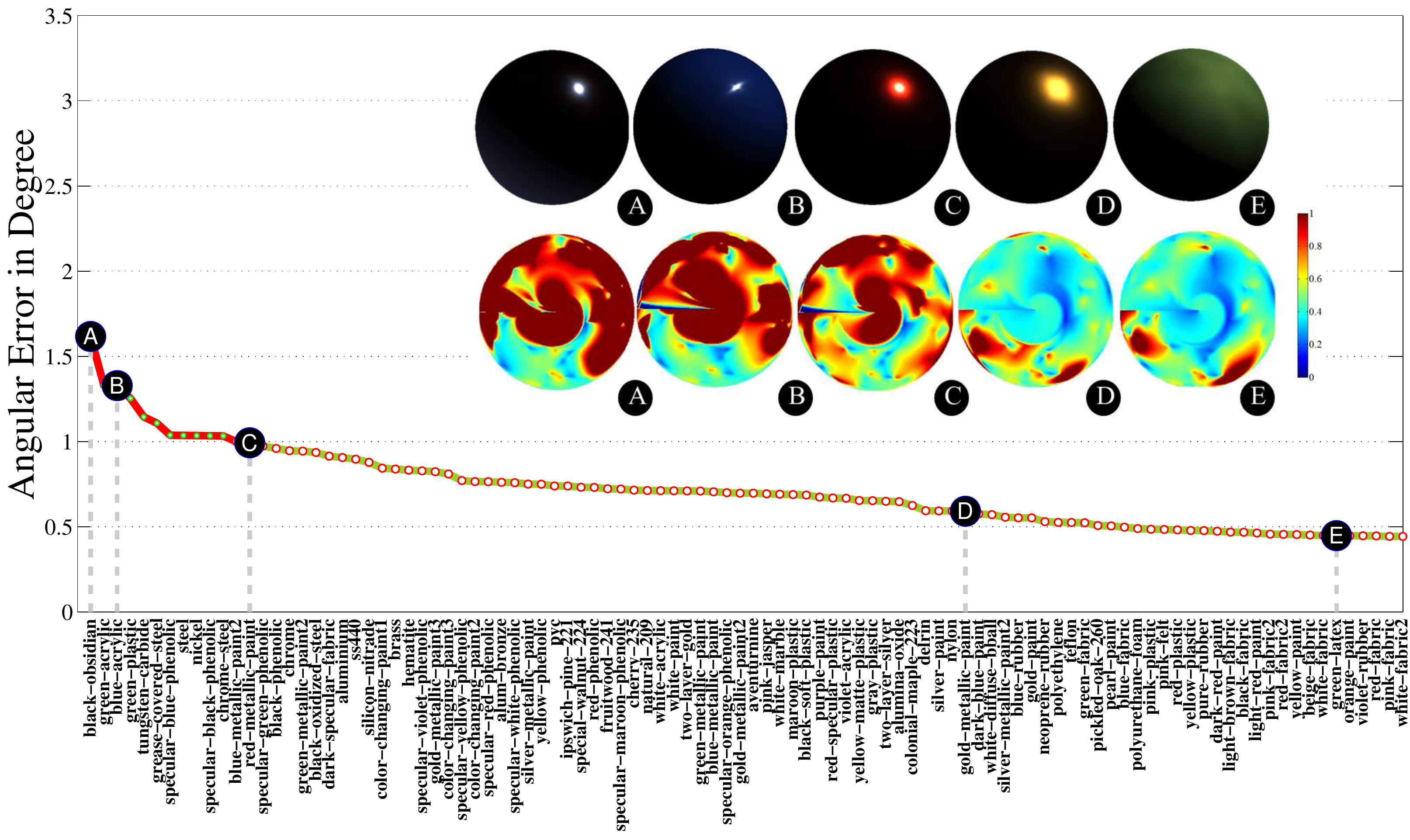}
	\caption{\textbf{Normal estimation for different materials.} We fix the number of input images/lighting directions to 253. For each material BRDF, we compute average error over 50,000 randomly-generated surface normals. Inset are the angular error distribution for a few select materials. }
	\label{fig:mean_err}
\end{figure}

\begin{figure*}
	\includegraphics[width=\textwidth]{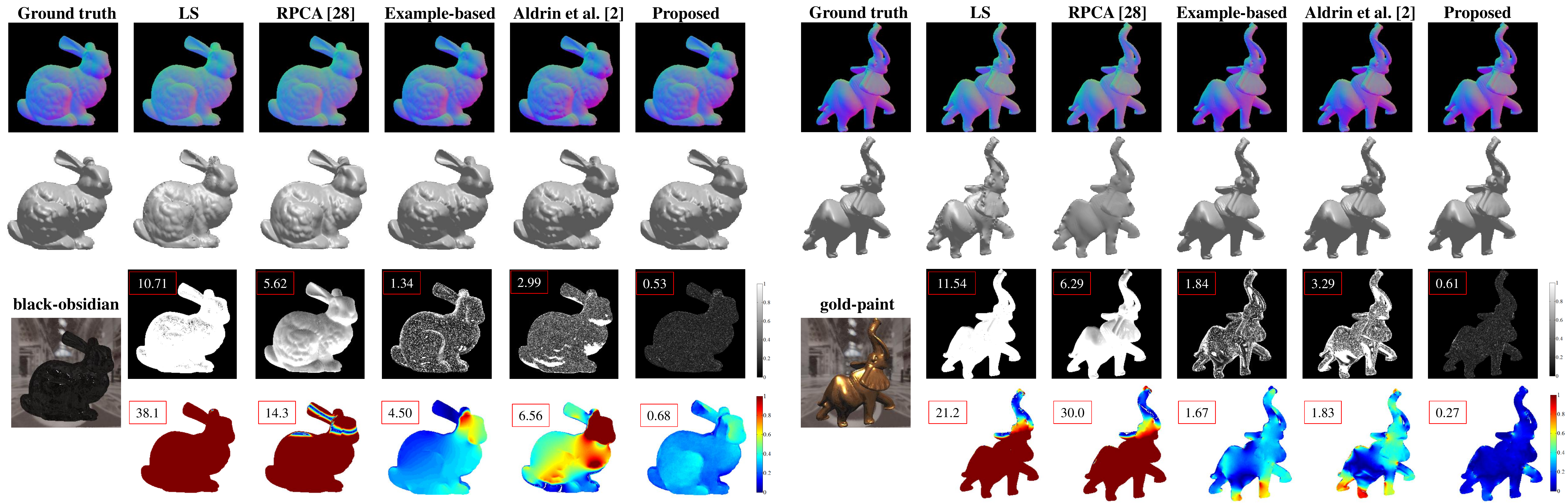}
	\caption{\textbf{Normal estimation across algorithms.}  We compare the performance of photometric stereo under Lambertian model (LS), robust PCA-based approach \cite{wu2011robust} (RPCA),  simulated example-based \cite{hertzmann2005example}, Alldrin et al.\ \cite{alldrin2008photometric} on two objects suing 253 images each. Shown are (top-bottom) the estimated surface normals, recovered 3D surface, angular error in normal estimation in degrees and relative error in depth map based on different approaches. The insets in rows 3 and 4 are the average normal errors in degrees and the relative depth errors.}
	\label{fig:bunny_err}
\end{figure*}

\vspace{-3mm}
\paragraph{Varying BRDF.}
In Figure \ref{fig:mean_err}, we evaluate performance of surface normal estimation for varying material BRDFs. 
We fixed the number of images at $Q = 253$.
Shown are aggregate statistics computed over 50,000 randomly generated surface normals.
The worst case error is less than $2^\circ$ and the error tapers down to $0.5^\circ$ which is the finest sampling that we used for generating candidate normals. 
This can presumably be reduced by either choosing a finer sampling grid or using gradient descent techniques.

\vspace{-3mm}
\paragraph{Comparisons.}
Figure \ref{fig:bunny_err} showcases the performance of many photometric stereo techniques for different objects: a {\tt black-obsidian} bunny and a {\tt gold-painted} elephant.
We used 253 input images for each object.
Photometric stereo under Lambertian model fails to recover precise normal maps indicating the presence of non-Lambertian components.
The robust PCA-based solver \cite{wu2011robust} produces better normal maps as compared to the traditional photometric stereo, however it produces overly smoothed estimates; this can be attributed to removal of non-Lambertian cues which are invaluable for precise normal estimation.
The solution of Alldrin et al.\ \cite{alldrin2008photometric} while significantly better than Lambertian technique produces errors greater than $1^\circ$.
In contrast, the proposed method returns reliable normal estimates for both scenes indicating the robustness of the underlying solution. 
We also simulated the performance of example-based photometric stereo which is identical to the proposed technique when applied to a two-material ({\tt white-diffuse} and {\tt chrome}) dictionary. As expected, having a larger dictionary of BRDFs as in the proposed technique does provide significant improvements in surface normal estimation.

%
\vspace{-3mm}
\paragraph{Performance of BRDF estimation.} 
Given a test BRDF, we generated $100$ surface normals with random orientations and rendered their appearance for $253$ lighting directions. 
Assuming the knowledge of the true surface normals, we estimate the BRDF using the optimization in Section \ref{sec:BRDF}.

\begin{figure}[!ttt]
	\includegraphics[width=.475\textwidth]{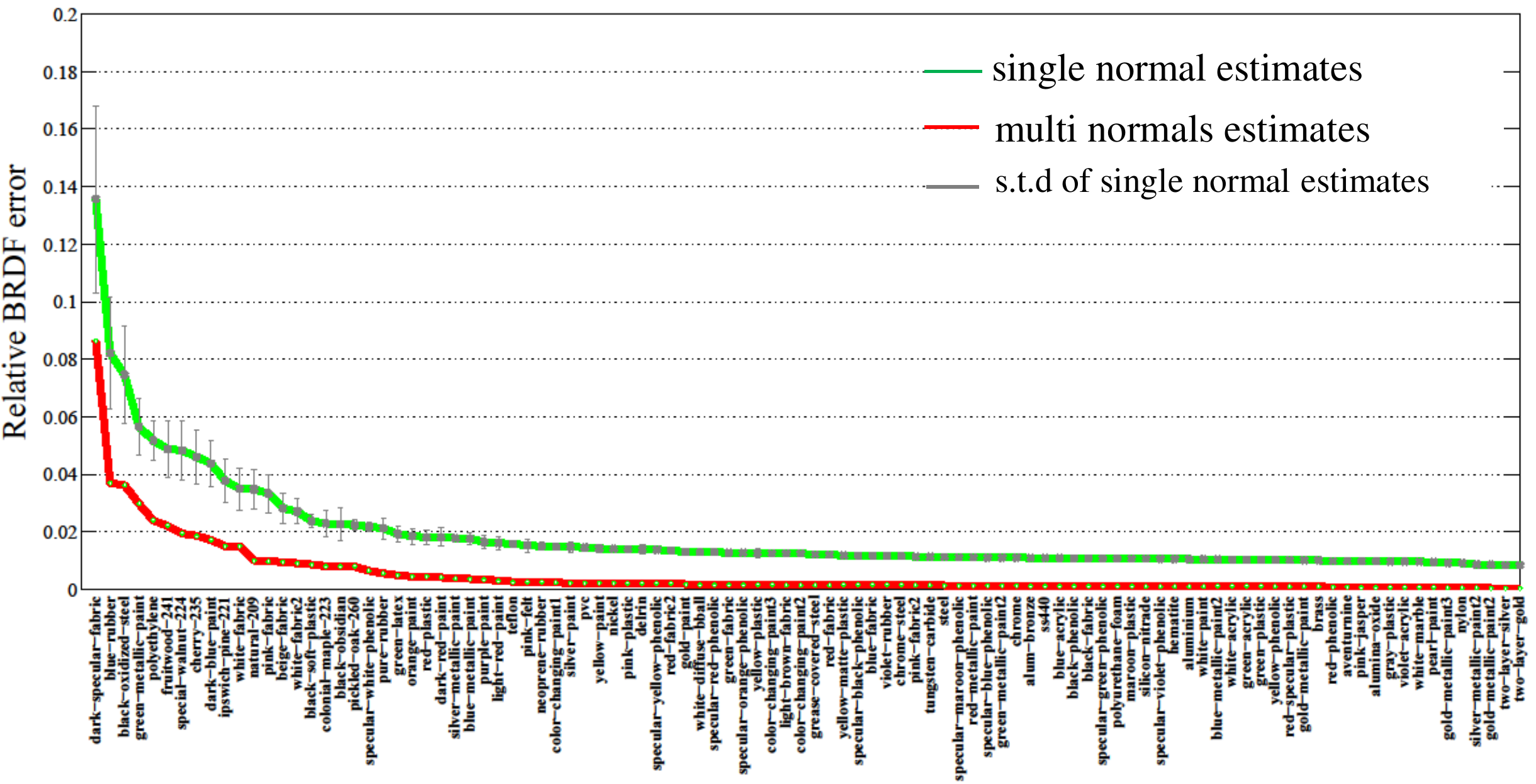}
	\caption{ \textbf{Quantitative BRDF evaluation.} Given $253$ lighting directions, we evaluate accuracy of BRDF estimation across different materials. For each material, we generated $100$ normals with random orientations and estimated the BRDF for each instance individually (per-pixel) as well as collectively. For the per-pixel estimates, we plot average and standard deviation of the errors.}
	\label{fig: brdf_err}
\end{figure}
\begin{figure}[!ttt]
	\includegraphics[width=.475\textwidth]{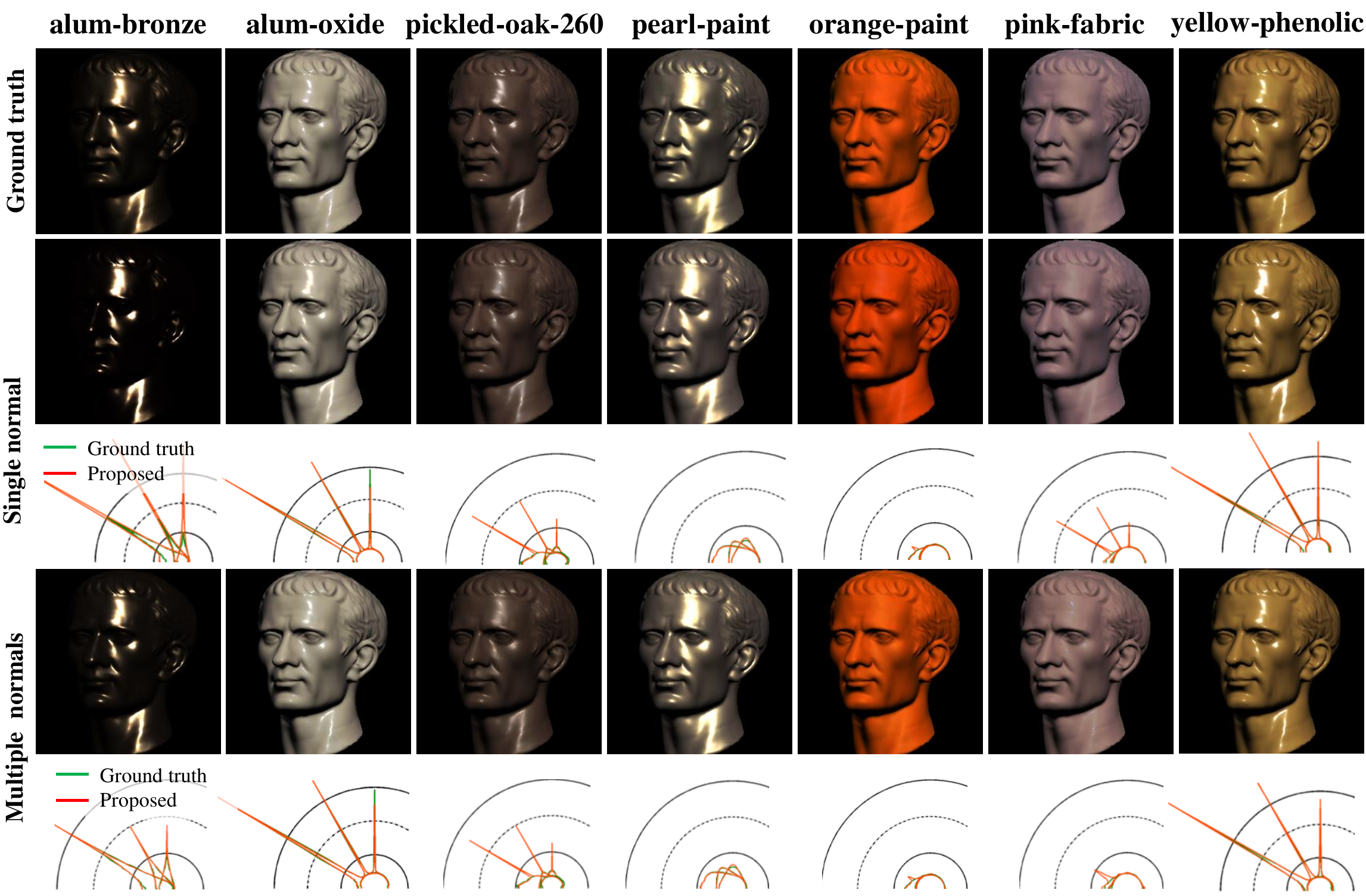}
	\caption{\textbf{Qualitative BRDF evalution.}  Shown are rendered BRDF for the \textit{Ceasar} statue for a few select material from MERL database \cite{matusik2003data}. (row 1) The rendered image based on ground truth BRDF; (rows 2 and 4) rendered images based on estimated BRDF from a single normal and 100 randomly generated normals, respectively; (rows 3 and 5) the polar plot for the reflectance function for the incident light angles [$0^\circ$, $30^\circ$, $60^\circ$]. }
	\label{fig: comp_brdf}
\end{figure}

We characterize the performance of the per-pixel BRDF estimate as well as the error in the BRDF estimate when the information at the 100 normals are pooled.
We use the relative BRDF error \cite{ngan2005experimental} to quantify the accuracy of the estimate.
Given true BRDF $\rho$ and estimated value $\widehat{\rho}$, the relative BRDF error is given as 
\begin{equation}
\sqrt{ {\sum_i w_i ((\widehat{\rho}(i)-\rho(i))\cdot \max(0, \cos(\theta_i)))^2}/{\sum_i w_i}}, 
\label{eqn:relBRDF}
\end{equation}
with $w_i$ set equal to $1$ for convenience.

Figure \ref{fig: brdf_err} shows the errors for different materials in the database --- rank-ordered from worst-to-best performance --- both for the per-pixel BRDF estimation as well as the joint estimation.
In Figure \ref{fig: comp_brdf}, we show relighted images and polar plots for a subset of materials.

\vspace{-5mm}
\subsection{Real data}
%
Real images are present a layer of difficulty well beyond simulations and introduce inter-reflections, sub-surface scattering,  cast shadows,  and imprecise light source localization.
We test the performance of our shape and BRDF recovery algorithm on a wide range of datasets.
Specifically, we use images from two  sources ---  the light stage data from \cite{einarsson2006relighting}, and the gourd from \cite{alldrin2008photometric}.
%

%

\begin{figure}[[!ttt]
	\includegraphics[width=.475\textwidth]{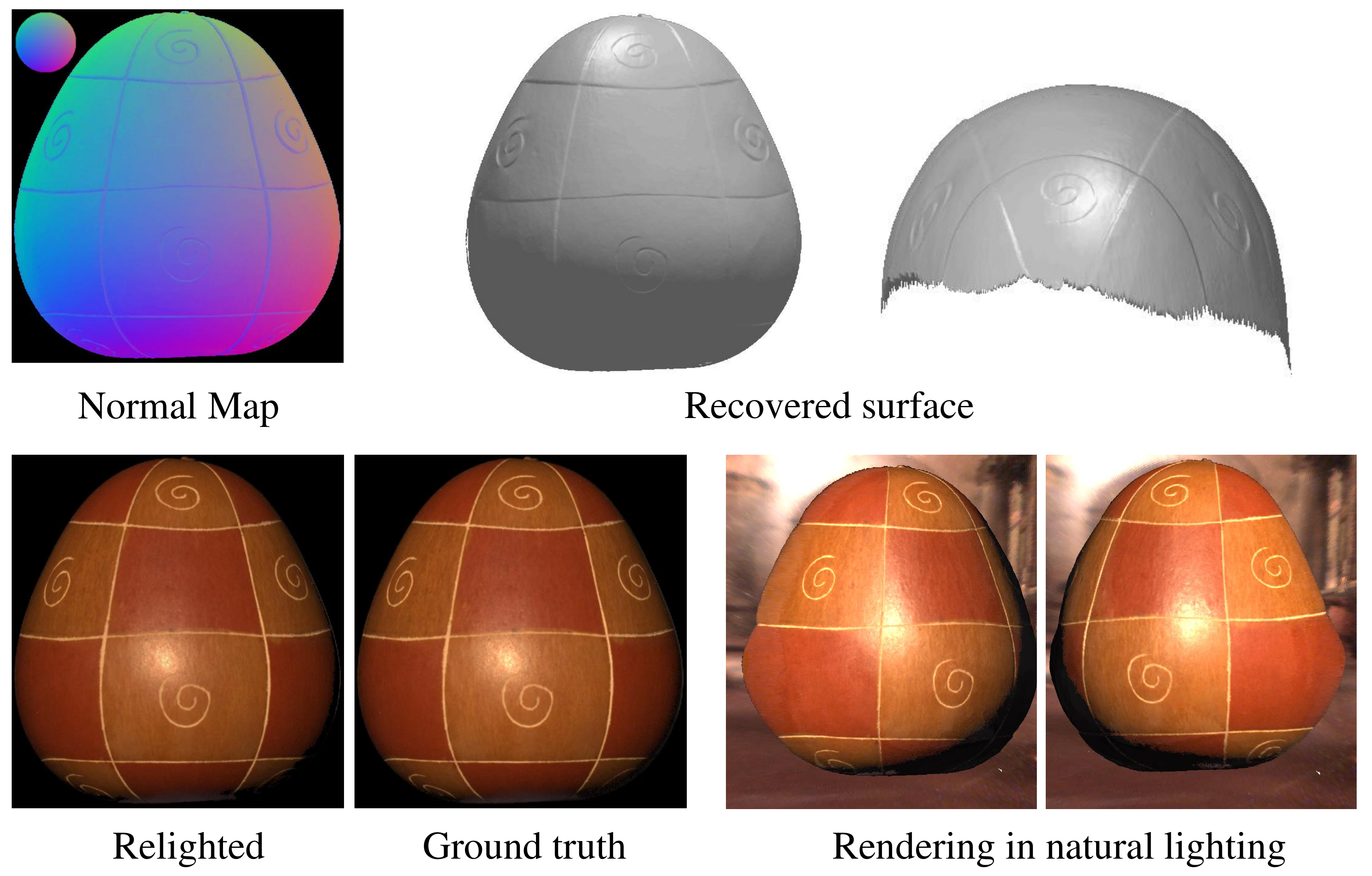}
	\caption{\textbf{Results on ``gourd1'' dataset.} We show the estimated normal map in false color (top-left) and 3D surface (top-right) recovered from it. We also show the relighting results (bottom-left), ground truth under the same lighting direction (bottom-middle), and relighting under natural environment (bottom-right).}
	\label{fig:gourd1}
\end{figure}

\begin{figure}[[!ttt]
	\includegraphics[width=.475\textwidth]{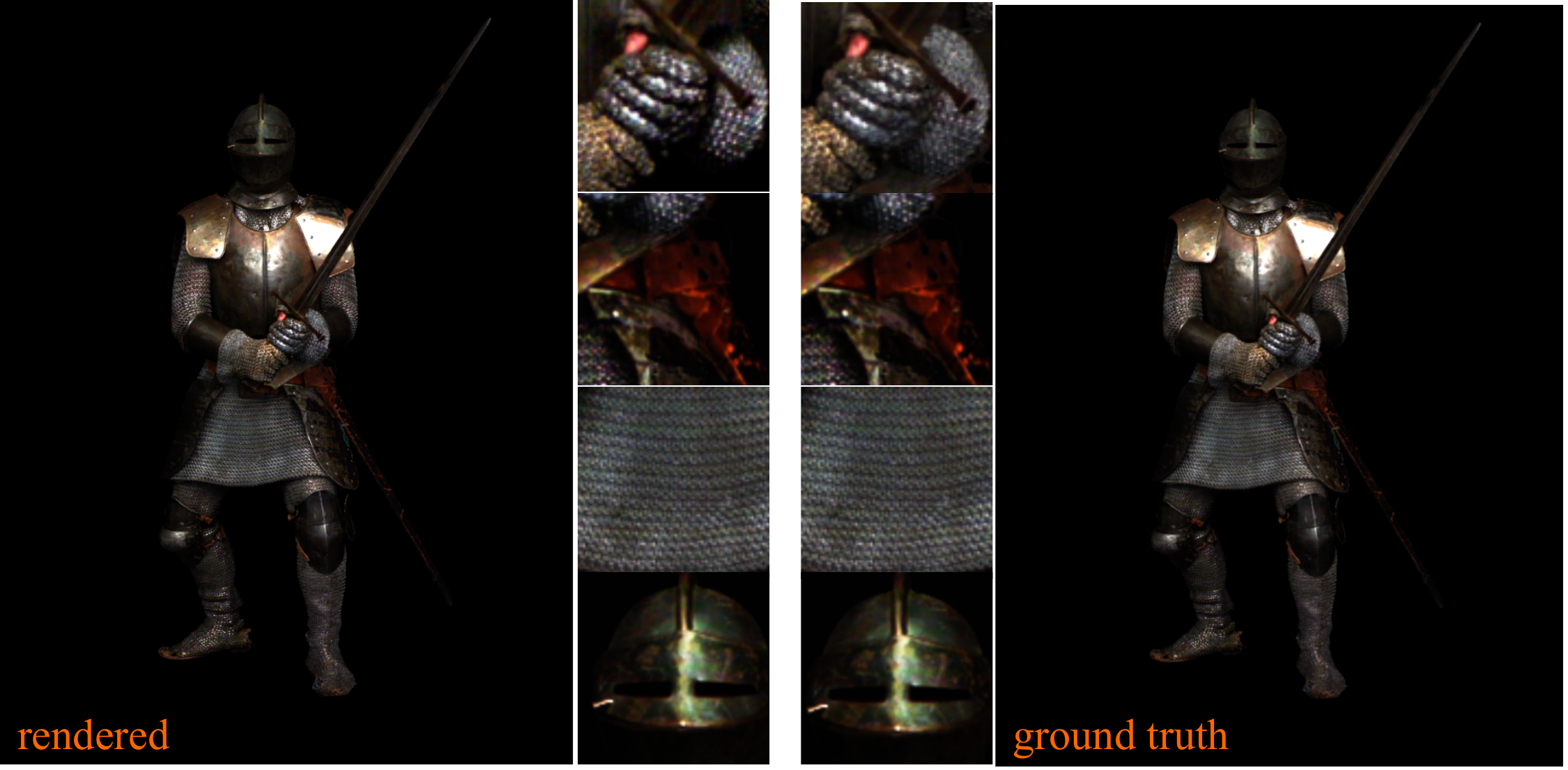}
	\caption{\textbf{Relighting results on ``knight\_fighting'' dataset.}}
	\label{fig:knight_fighting}
\end{figure}

\begin{figure*}
\center
	\includegraphics[width=\textwidth]{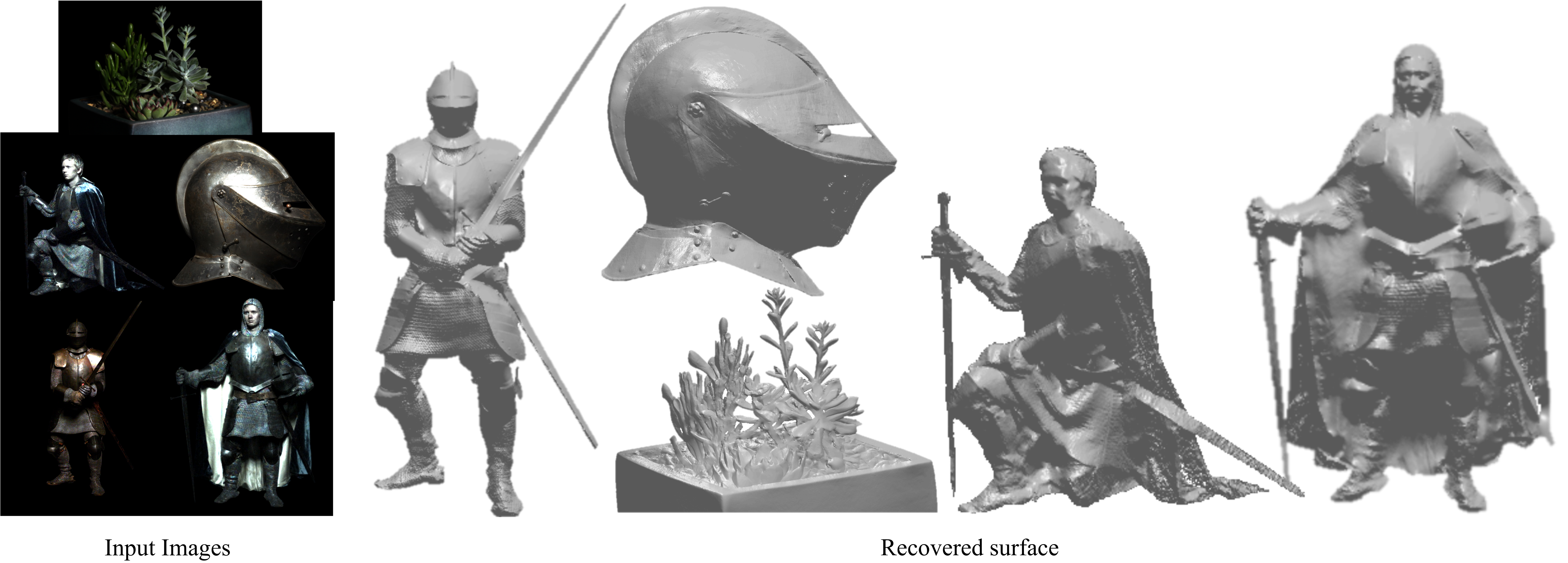}
	\caption{\textbf{Recovered surfaces on several real scenes with complex, spatially varying reflectance.}}
	\label{fig:shape_estimate}
\end{figure*}

%
Figures~\ref{fig: teaser}, \ref{fig:gourd1}, \ref{fig:knight_fighting} and \ref{fig:shape_estimate} showcase the performance of our algorithm on the real datasets.
The results in Figure~\ref{fig: teaser}, \ref{fig:knight_fighting} and \ref{fig:shape_estimate} were obtained from 250 input images, and the results of ``gourd1'' in Figure \ref{fig:gourd1} was obtained from 100 input images.
The recovered shape and BRDF (as visualized via rendered images) seem to be in agreement with the results  in \cite{alldrin2008photometric}; however, our algorithm is significantly simpler and employs a per-pixel algorithm that be easily parallelized.

The robustness of the per-pixel BRDF estimate is tested in Figure \ref{fig:knight_fighting} where there are not just a wide variety of unique materials (the helmet, the breast-plate, the chain, the red scabbard, to name a few) but also significant modeling deviations (inter-reflections, cast-shadows).
In spite of this, our approach produces a faithful rendition of the scene.
The per-pixel BRDF estimation allows us to handle objects with  complex spatial variations.
In contrast, methods that assume the presence of just a few reference BRDFs as in \cite{goldman2010shape, alldrin2008photometric} would not scale easily to such scenes.
We refer the reader to the supplemental  videos highlighting the relighting results.

\section{Discussions} \label{sec:discuss}
We present a photometric stereo technique for per-pixel normal and BRDF estimation for objects that are visually complex.
We demonstrate that the use of a BRDF dictionary significantly simplifies the inverse problem and provides not just state-of-the-art results in normal and BRDF estimation but also works robustly on a wide range of real scenes.
A hallmark of our approach is the ability to obtain per-pixel BRDF estimation without any spatial smoothness assumptions endemic to state-of-the-art SV-BRDF estimation techniques \cite{goldman2010shape,alldrin2008photometric}; this makes it applicable to scenes with a large number of unique materials.
Finally, our per-pixel framework is ripe for further speed-ups by solve for the shape and reflectance at each pixel in parallel.

\vspace{-3mm}
\paragraph{Limitations.} 
While the use of virtual examples provides flexibility beyond \cite{hertzmann2005example}, we require light calibration and hence, our method is most suited to shape and reflectance acquisition from light-stages where the light sources are fixed and the calibration is a one-time effort.
The accuracy of our coarse-to-fine normal estimation is lower bounded by finest sampling of our candidate normals.
This can potentially be improved by  refine the estimates using a gradient descent scheme starting with our solution; however, this approach could be computationally intensive.
The SV-BRDF produced by our approach can be noisy especially since we independently recover the BRDF at each pixel.
If we have a priori knowledge that the scene has a limited number of unique materials, then enforcing this could lead to robust SV-BRDF estimates.
This can be easily incorporated into our framework by enforcing the matrix of  sparse coefficients $[ \bfc_1 \ldots \bfc_\bfp  \ldots]$ to be low-rank.
Finally, it is also important that the scene lies in the linear span of our dictionary. In the failure of this, our results can be unpredictable. 
Here, the need for a larger dictionary encompassing hundreds, if not thousands, of materials  would be invaluable for the broader applicability of our method.

{\small
\bibliographystyle{ieee}
\bibliography{refs}
}

\end{document}